\pdfoutput=1
\documentclass{article}

\usepackage[preprint]{neurips_2023}

\usepackage{graphicx}
\usepackage{subcaption}

\usepackage[utf8]{inputenc} 
\usepackage[T1]{fontenc}    
\usepackage{hyperref}       
\usepackage{url}            
\usepackage{booktabs}       
\usepackage{amsfonts}       
\usepackage{nicefrac}       
\usepackage{microtype}      
\usepackage{xcolor}         
\newcommand{\nosection}[1]{\vspace{2pt}\noindent\textbf{#1.}}
\usepackage[most]{tcolorbox}

\title{Is GPT4 a Good Trader?}

\author{
  BingzheWu \\
  Tencent AI Lab\\
}

\begin{document}

\maketitle

\section{Introduction}

\emph{Speculation is as old as the hills because human nature never changes.---Jesse Livermore}

The realm of stock market analysis has witnessed a myriad of evolutions over the past century, with a distinctive drift from rudimentary practices to a more systematized analytical approach known as technical analysis \cite{rhea2013dow,schannep2008dow,rhea1993dow,hyerczyk2009pattern}. The essence of technical analysis revolves around the examination of past market data, primarily price and volume, to prognosticate future market movements. Pioneering this analytical trajectory, Dow Theory emerged in the late 19th century, premised on the assertion that markets move in identifiable patterns and trends that are reflective of the collective sentiment of the entire market \cite{rhea1993dow}. Almost concomitantly, the Gann angles or Gann Theory\cite{hyerczyk2009pattern}, posited by William Delbert Gann, fortified the edifice of technical analysis by elucidating the relationship between price and time through geometric patterns.

Yet, as formidable as these theories are, they present a slew of challenges when applied pragmatically by investors. Foremost, they often lack quantifiable metrics that can be consistently employed across diverse market conditions. The theoretical constructs, while enlightening, are frequently deemed as arcane, leading to interpretational challenges even among seasoned investors. Additionally, the absence of a unified standard when applying these theories across disparate trading scenarios exacerbates the inherent unpredictability of stock market behaviors. Such challenges underscore the imperative for continuous refinement in the domain of technical analysis, ensuring its congruence with contemporary market intricacies and the evolving digital landscape \cite{rhea2013dow}.

Recently, large language models (LLMs), particularly GPT-4, have demonstrated significant capabilities in various planning and reasoning tasks \cite{cheng2023gpt4,bubeck2023sparks}. Motivated by these advancements, there has been a surge of interest among researchers to harness the capabilities of GPT-4 for the automated design of quantitative factors that do not overlap with existing factor libraries, with an aspiration to achieve alpha returns \cite{webpagequant}. In contrast to these work, this study aims to examine the fidelity of GPT-4's comprehension of classic trading theories and its proficiency in applying its code interpreter abilities to real-world trading data analysis. Such an exploration is instrumental in discerning whether the underlying logic GPT-4 employs for trading is intrinsically reliable. Furthermore, given the acknowledged interpretative latitude inherent in most trading theories, we seek to distill more precise methodologies of deploying these theories from GPT-4's analytical process, potentially offering invaluable insights to human traders. 

To achieve this objective, we selected daily candlestick (K-line) data from specific periods for certain assets, such as the Shanghai Stock Index. Through meticulous prompt engineering, we guided GPT-4 to analyze the technical structures embedded within this data, based on specific theories like the Elliott Wave Theory. We then subjected its analytical output to manual evaluation, assessing its interpretative depth and accuracy vis-à-vis these trading theories from multiple dimensions. The results and findings from this study could pave the way for a synergistic amalgamation of human expertise and AI-driven insights in the realm of trading.

\section{Case Study}

For our case study, we focus on the Shanghai Stock Index from March 2020 to February 2021. We aim to assess GPT-4's competency in technical analysis, specifically in terms of basic technical skills, Elliott Wave Theory, and Dow Theory. Each dimension will be elucidated by detailing the task setting, prompt engineering, and subsequent analysis of the results.




\subsection{Elliott Wave Theory}
The Elliot Wave Theory \cite{wave}, postulated by Ralph Nelson Elliott in the 1930s, posits that collective investor psychology, or crowd psychology, moves between optimism and pessimism in natural sequences. These mood swings create specific patterns, manifested in market price movements, which are fractal in nature. The primary pattern consists of five waves, often described as follows: (1) Three impulse waves (labeled 1, 3, and 5) that follow the main trend. (2)Two corrective waves (labeled 2 and 4) that move against the trend.

The impulse waves typically embody a strong force in the direction of the prevailing trend, whereas the corrective waves signify a weaker movement against it.
This part sought to evaluate how accurately GPT-4 can discern and label the various waves within the Elliott Wave structure on the Shanghai Stock Index(00001.SH) during the study period. Our evaluation criteria include the model's foundational comprehension, clarity in the analytical process, the validity of the final conclusions, and a broader assessment of GPT-4's technical analytical proficiency.

\nosection{Task Description and Prompt }
The task is to identify potential five-Wave structure in the given index data. In practice, we have found that GPT-4, when not provided with initial constraints, is highly likely to generate code that lacks logical coherence. Therefore, we incorporate additional constraints in the prompts to ensure a more controlled and coherent output. Readers are encouraged to experiment with additional constraints and prior knowledge to achieve even better results.
The basic prompt we used in this test is as follows:

\begin{tcolorbox}[colback=gray!30!white,colframe=blue!50!white, title=Prompt for Elliott Wave Theory Test]

\textcolor{red}{Role: }You are a trader on Wall Street, responsible for foreign exchange, commodities, stocks, and other types of trades. You are proficient in all the famous technical trading frameworks such as the Elliott Wave Theory and Dow Theory. \\
\textcolor{red}{Objective: }Analyze the technical trend characteristics of the corresponding subject based on the K-line data uploaded by the user.\\
\textcolor{red}{Task Description:}\\
1. Analyze its five-wave structure using the Elliott Wave Theory based on the K-line data uploaded by the user.\\
2. Output the start and end dates of each wave segment.\\
3. Provide detailed deductive steps.\\
\textcolor{red}{Init condition: }\\
1. The wave1 starts from 20200320.
\end{tcolorbox}

\begin{figure}
    \centering
    \begin{subfigure}[b]{0.45\textwidth}
        \includegraphics[width=\linewidth]{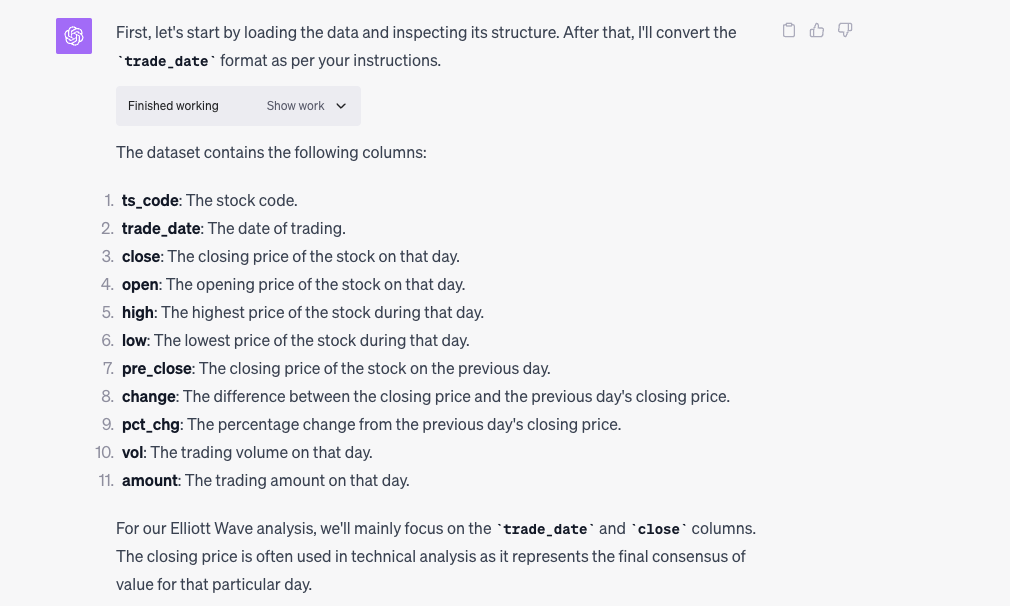}
        \caption{Data Process.}
    \end{subfigure}
    \hfill
    \begin{subfigure}[b]{0.45\textwidth}
        \includegraphics[width=\linewidth]{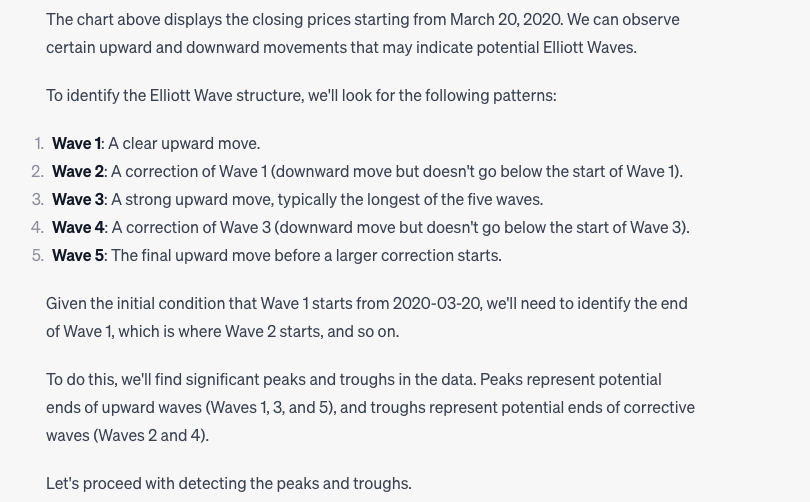}
        \caption{Knowledge recall and task planning.}
    \end{subfigure}

    \begin{subfigure}[b]{0.45\textwidth}
        \includegraphics[width=\linewidth]{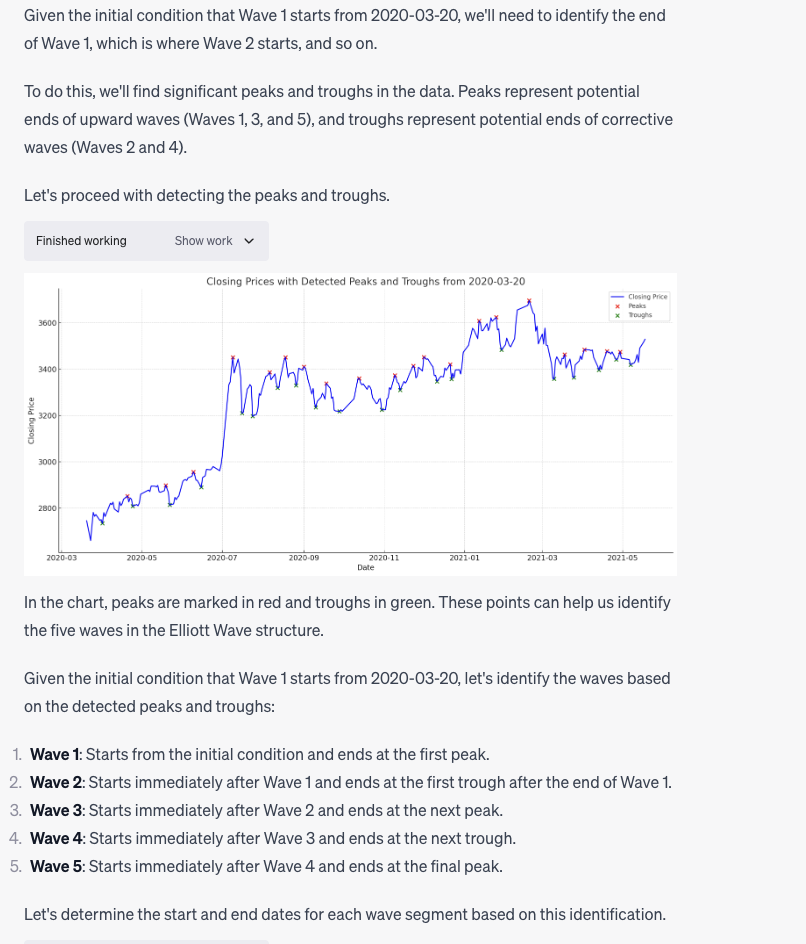}
        \caption{Step action.}
    \end{subfigure}
    \hfill
    \begin{subfigure}[b]{0.45\textwidth}
        \includegraphics[width=\linewidth]{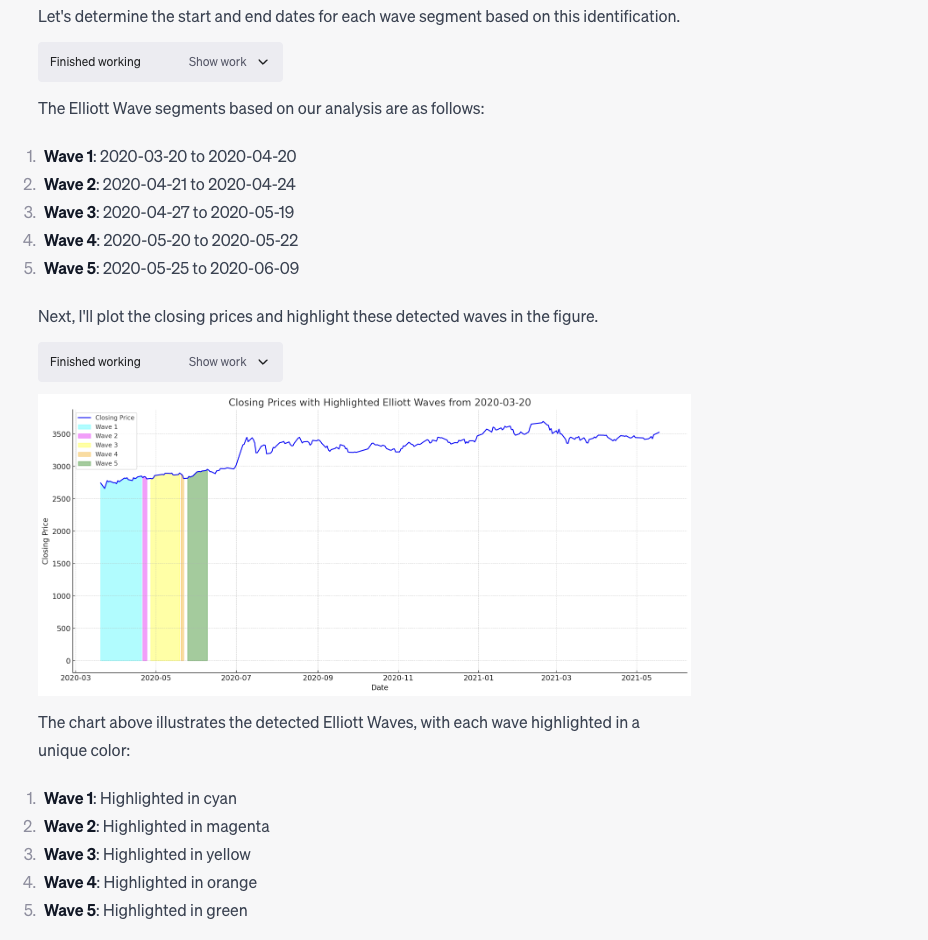}
        \caption{Final results.}
    \end{subfigure}
    \caption{Overall process.}\label{fig:wave_overall}
\end{figure}

\nosection{Evaluation}
The overall analysis of GPT4 can be divided into the following phase:
\begin{enumerate}
    \item Data process: Loading user-uploaded data into a specific format and generating visualizations is a critical process in data analysis and visualization tasks.
    \item Knowledge recall and task planning: 
Retrieving background knowledge pertaining to the recall task (in this context, the theory of waves) and formulating a comprehensive plan for the task are crucial initial steps.
    \item Action:According to the task plan, invoking the code interpreter is performed to execute various segmented steps.
    \item Result output:Integrate all the intermediate findings as elucidated above and subsequently present the ultimate analytical outcomes, accompanied by pertinent graphical representations.
\end{enumerate}
Figure \ref{fig:wave_overall} illustrates select pivotal intermediate results.
Next, we will assess the plausibility of GPT-4 output at each step of the evaluation process. We will construct a checklist and distribute it to five experienced traders for evaluation. Currently, each step consists of a five-level scoring system, namely: 
\begin{itemize}
    \item Completely unreasonable(0) 
    \item Barely relevant(0.25)
    \item Marginally acceptable(0.5) 
    \item Good (0.75)
    \item Great, inspiring response(1)
\end{itemize}
In our subsequent work, we will attempt to develop more granular metrics. After completing the scoring process, we will calculate the average score for each dimension to obtain the final score.

The comprehensive quantitative results are presented in Table 1, indicating that GPT-4 achieves commendable scores in both task planning and subtask execution. However, its overall score experiences a sudden drop to $0.15$ in the final result. A detailed examination of each step reveals that, as demonstrated in Figure 1b, GPT-4 can accurately recall the core principles of wave theory and eloquently present the technical characteristics of each wave. In response to these features, GPT-4 attempts to identify local extrema and corresponding substructures. It is evident that this step effectively captures the core task and successfully executes the planning phase.

Regarding the execution of subtasks, GPT-4 largely adheres to the earlier planning stages. However, in the final step, we observe that the five-wave structure identified by GPT-4 is primarily based on local trends, whereas professional traders typically prioritize analyzing the overall trends of the asset when applying wave theory. Based on this observation, we conducted interactive corrective prompts to encourage GPT-4 to adopt a global perspective.

\begin{table}[h]
\centering
\begin{tabular}{|c|c|c|c|}
\hline
Prompting Strategy &Knowledge recall \& task planning   & Sub-task Quality& Final Results \\
\hline
Original & 0.84 & 0.8 & 0.15 \\
\hline
Expertise injection & 0.84 &0.8& 0.25 \\
\hline
Expertise+correction & 0.84 & 0.8& 0.45 \\
\hline
\end{tabular}
\caption{Step-wise human evaluation.}
\end{table}
\begin{figure}[http]
    \centering
    
    \begin{subfigure}{0.45\textwidth}
        \includegraphics[width=\linewidth]{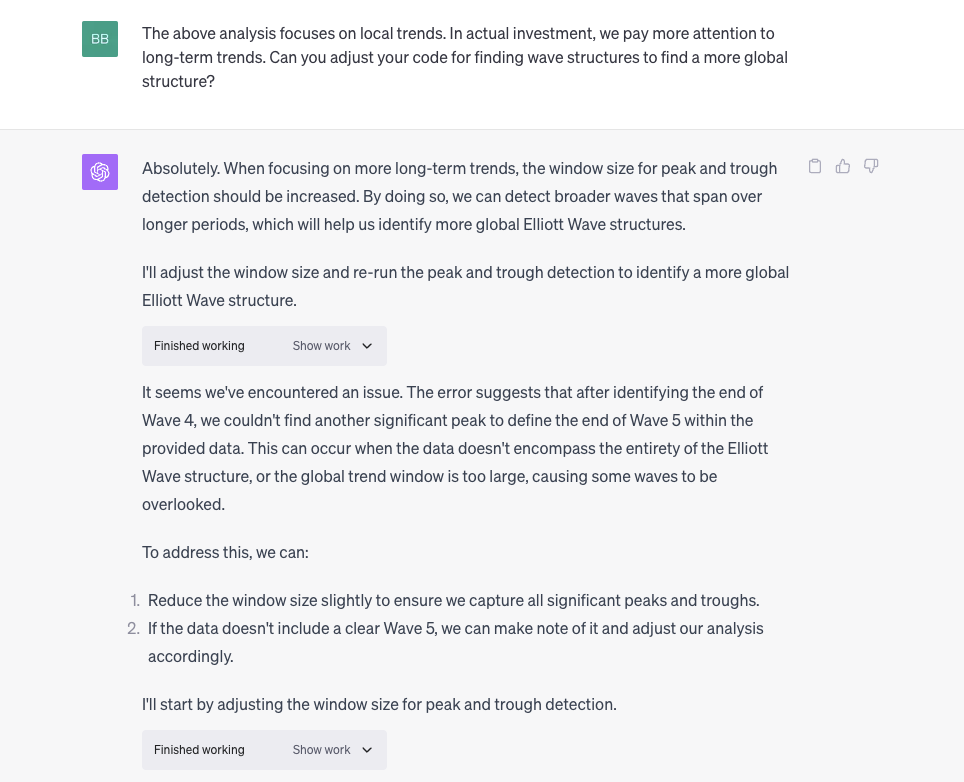}
    \end{subfigure}
    \hfill
    \begin{subfigure}{0.45\textwidth}
        \includegraphics[width=\linewidth]{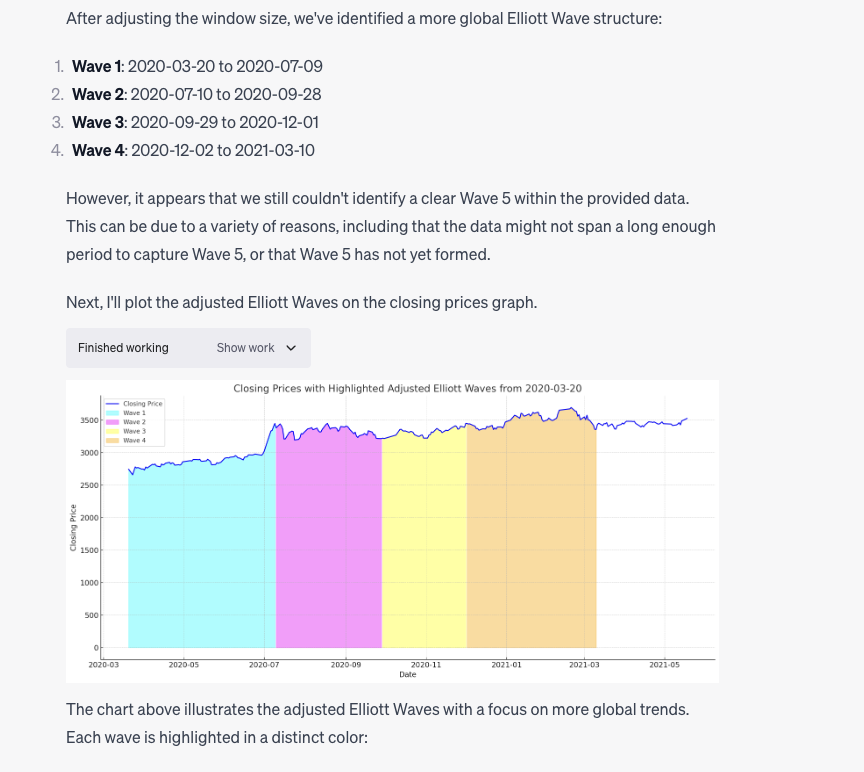}
    \end{subfigure}
    \caption{By injecting post-trader preference prompts and the reasoning process into the system.}
    
    \vskip\baselineskip
    
    \begin{subfigure}{0.45\textwidth}
        \includegraphics[width=\linewidth]{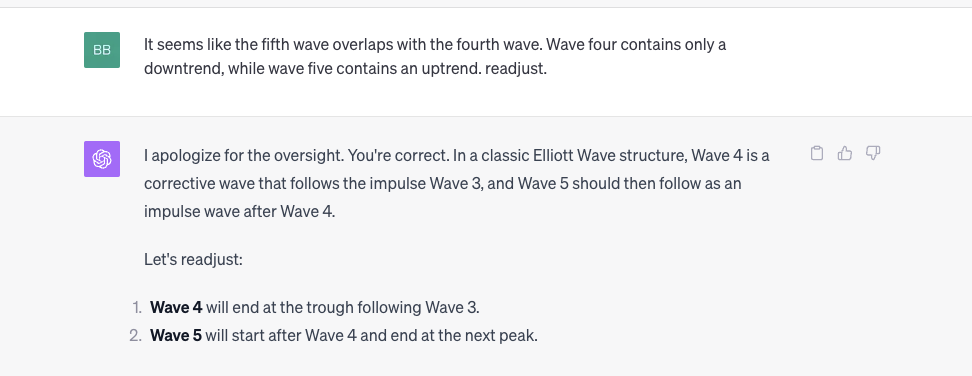}
    \end{subfigure}
    \hfill
    \begin{subfigure}{0.45\textwidth}
        \includegraphics[width=\linewidth]{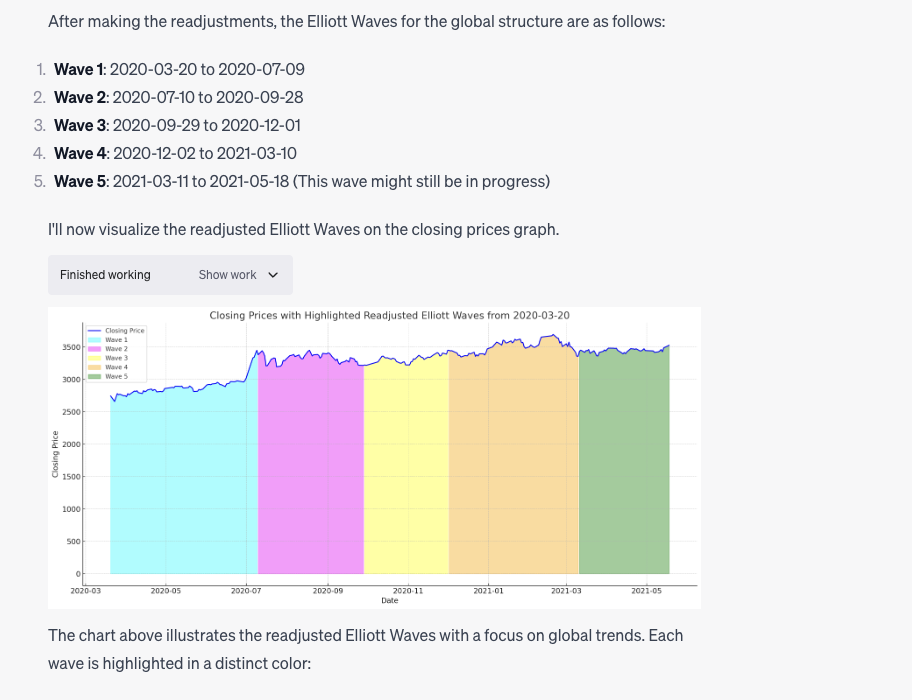}
    \end{subfigure}
    \caption{The prompt and outcomes that involve corrections to intermediate processes are subject to translation and refinement.}
    
\end{figure}

By interactively modifying the prompts to inject the trader's preferences into GPT-4's reasoning process, as illustrated in Figure 2, we can observe that GPT-4, upon receiving new prompts, swiftly corrects its approach to identifying wave patterns and shifts towards a more global perspective in search of wave structures. However, in this particular attempt, it overlooks the structure of the fifth wave and erroneously introduces a portion of a mid-term upward trend within the fourth wave, which contradicts the principles of wave theory. Consequently, we further guide its correction of this error (Figure 3), ultimately achieving a reasonably improved score, although the fourth wave still does not adhere to the principles of wave theory.

From the detailed process outlined above, it becomes evident that, despite the capacity of GPT-4 to comprehensively grasp the theory of waves from textual inputs, practical applications reveal a misalignment between GPT-4 and traders' preferences, primarily attributed to the flexibility inherent in the theory of waves. This observation gives rise to an intriguing avenue for research: while techniques such as Reinforcement Learning from Human Feedback (RLHF) have demonstrated considerable success in aligning textual prompts with human preferences at the level of values and instructions, addressing the long-tail human preferences in specific domains and bridging the gap between textual-level preferences and preferences manifested in tool usage and code generation remains a formidable challenge for GPT-4. Resolving this challenge is pivotal for the effective deployment of large language models in real-world tasks.

\subsection{Dow Theory}
To be continued.

\section{Conclusion and Future Work}
From the case study presented in this article, it is evident that GPT-4 demonstrates a reasonably accurate comprehension of textual aspects within the domain of trading theory. However, there remains a substantial distance to cover before these theoretical understandings can be effectively applied to real-world data. In light of the discussions within this paper, several unexplored research directions emerge, which are as follows:
\begin{itemize}
    \item Transition from Textual Alignment to Tool Utilization and Code Generation Alignment: A key challenge lies in bridging the gap between understanding textual instructions and aligning them with the practical use of tools and code generation.
    \item Dynamic Injection of Domain Knowledge during GPT-4 Inference: The ability to dynamically inject domain-specific understanding into GPT-4's reasoning process interactively is an area that warrants exploration.
    \item Beyond Basic Trading Theory: While this paper primarily focuses on evaluating GPT-4's comprehension of fundamental trading theories, it is equally important to consider how this model can be leveraged to discover novel trading signals (factors). 
\end{itemize}

In our future work, we intend to expand our research efforts in several directions. Firstly, we plan to assess GPT-4's understanding of more intricate and complex trading theories, including esoteric ones like  Chinese entanglement theory\cite{}. Additionally, we are exploring the fine-tuning of trading agents based on open-source models in the 13B-70B parameter range. This endeavor will be complemented by integrating open-code interpreters to facilitate local analysis of trading data by end-users.
\bibliographystyle{plainnat}
\bibliography{ref}
\end{document}